\documentclass[10pt,twocolumn,letterpaper]{article}

\usepackage[pagenumbers]{cvpr} 

\usepackage{graphicx}
\usepackage{amsmath}
\usepackage{amssymb}
\usepackage{booktabs}

\usepackage[pagebackref,breaklinks,colorlinks]{hyperref}

\usepackage[capitalize]{cleveref}
\crefname{section}{Sec.}{Secs.}
\Crefname{section}{Section}{Sections}
\Crefname{table}{Table}{Tables}
\crefname{table}{Tab.}{Tabs.}

\def\cvprPaperID{*****} 
\def\confName{CVPR}
\def\confYear{2022}

\begin{document}

\title{Dog nose print matching with dual global descriptor based on Contrastive Learning}

\author{Bin Li\\
Yunnan University\\
{\tt\small flyingsheep@mail.ynu.edu.cn}
\and
Zhongan Wang\\
ShanghaiTech University\\
{\tt\small wangzha@shanghaitech.edu.cn}
\and
Nan Wu\\
Yunnan University\\
{\tt\small deepfaker@mail.ynu.edu.cn}
\and
Shuai Shi\\
ShanghaiTech University\\
{\tt\small shishuai@shanghaitech.edu.cn}
\and
Qijun Ma\\
Zhejiang A\&F University\\
{\tt\small skypow2012@gmail.com}
}

\maketitle

\begin{abstract}
Recent studies in biometric-based identification tasks have shown that deep learning methods can achieve better performance.  These methods generally extract the global features as descriptor to represent the original image.  Nonetheless, it does not perform well for biometric identification under fine-grained tasks. The main reason is that the single image descriptor contains insufficient information to represent image. In this paper, we present a dual global descriptor model, which combines multiple global descriptors to exploit multi level image features. Moreover, we utilize a contrastive loss to enlarge the distance between image representations of confusing classes. The proposed framework achieves the top2 on the CVPR2022 Biometrics Workshop Pet Biometric Challenge. The source code and trained models are publicly available at: https://github.com/flyingsheepbin/pet-biometrics

\end{abstract}

\section{Introduction}
\label{sec:intro}

\begin{figure}[t]
  \centering
   \includegraphics[width=0.8\linewidth]{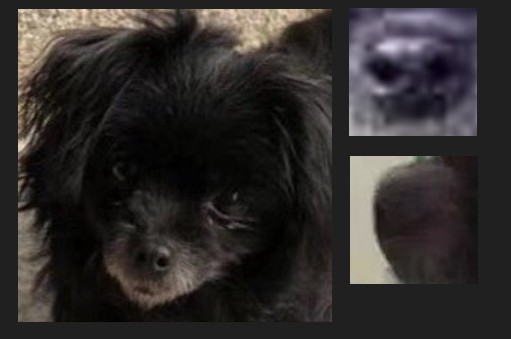}
   \caption{Example of nose print by dog Lian Lian.}
   \label{dog_nose}
\end{figure}

Vision-based pattern identification (such as the face, fingerprint, iris, etc.) has been successfully applied in human biometrics for a long history, however, there are few works that transfer these technologies to pet biometrics. Dog nose print matching aims to identify whether two dog nose print images belong to the same dog, as shown in Figure \ref{dog_nose}.

Ideally, the image of a dog's nose print is clearer and has a higher resolution, allowing you to see the texture of the image in detail. In practice, however, images of a dog's nose tend to be of low resolution, and the motion of a moving object can be blurred. 

In this paper, inspired by contrastive learning\cite{khosla2020supervised} and image retrieval tasks, we develop a method to solve these problems. Firstly, we solve the image blurred problem by contrasting dog nose print images with positive images and negative images, which helps the model to learn the similarity between positive pair images and differ the unlikely part between negative pair images. On the other hand, to solve the small image size that exists in the practice environment, we augmented the images during the training stage by resizing images to a smaller size with a probability. Extensive experiments show that our method can solve the above two problems.

The contributions of our proposed framework are as follows:
\begin{itemize}
\item To solve the dog nose print matching task, we designed a new model  with dual global descriptor, which can exploit the multi-scaled features of image.
\item In the dog nose print matching task, we firstly apply supervised contrastive learning to identify undistinguished sample.
\item Aiming at the problem of low image resolution, we apply the contrastive learning method and resetting image size method to alleviate the problem of blurred images effectively.
\end{itemize}

\begin{figure*}[h]
  \centering
   \includegraphics[width=0.8\linewidth]{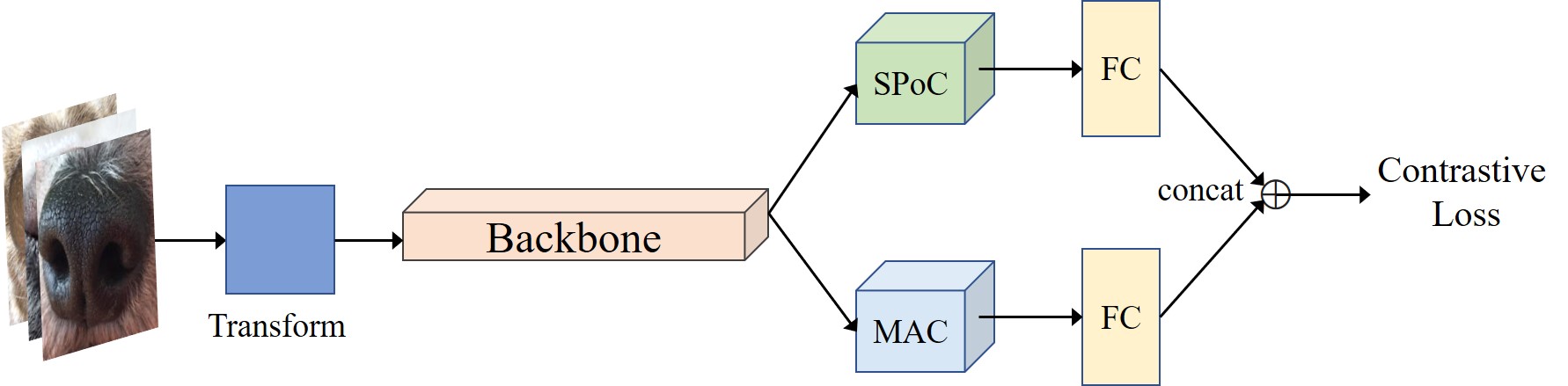}

   \caption{The architecture of our model with dual global descriptor based on Contrastive Learning.}
   \label{model_img}
\end{figure*}

\section{Related Work}
\label{sec:related}

\noindent

Animals can be distinguished by using physical characteristics, such as how humans have distinct fingerprint and iris patterns. In the past, animals were identified individually by animal tags. Not only does this cause mental and physical damage to the animal, it is also prone to duplication and forgery. Hence, biometric-based identification technology become more approved as alternatives to individual animal identification method. Handcrafted feature-based methods were utilized for identification. Kumar et al.\cite{kumar2016face} proposed a method to identify cattle using face images. The method extracted face features by using conventional machine learning algorithm. Chen et al.\cite{chen2016locality} introduced an identification method by using nose-print images with support vector method(SVM). Chakraborty et al.\cite{chakraborty2020investigation} used cropped muzzle images of pigs for breed identification. However, handcrafted feature-based approaches are more dependent on the quality of the dataset and prior knowledge, and cannot maintain high performance under a complicated environment. Recently, with the development of deep learning in computer vision, deep learning methods were widely used in image classification and object detection. Consequently, biometric-based identification based on deep learning has gradually attracted attention. Deb et al.\cite{deb2018face} introduced the PrimNet, a face recognition system, which can obtain images of three primates in the wild. Wang et al.\cite{wang2019learning} presented a residual CNN network to for gender classification by using the facial features of pandas. Han \cite{caya2021dog} proposed a DNNet method to an individual dog's nose-print pattern.

\section{Model}

Our model with dual global descriptor, which is based on contrast learning, is inspired by the combination of multiple global descriptors (CGD). The CGD framework has been demonstrated that it is flexible and expandable by global descriptors, backbones, losses and datasets. As illustrated in Figure 2, the architecture of our model is primarily divided into backbone and head. On the backbone side, it is straightforward to select the pretrained visual model as the feature extractor. On the head side, we used two types of the global descriptors, including SPoC descriptor based on sum-pooled convolutional features and MAC descriptor based on maximum activations of convolutions. Given an input of 3D tensor $V^{C\times H\times W}$, we made $V_{c}$ be the set of $H \times W$ activations for feature maps, where $C$ is the number of feature maps and $c\in\{1,...,C\}$. The vector $f$ as output, which is produced by global descriptor, can be generalized as follows:
\begin{align}
    f=[f_1,...,f_c,...,f_C]^T, f_c=(\frac{1}{|V_c|}\sum_{v\in V_c}v^{p_c})^{\frac{1}{p_c}},
\end{align}
where $p_c=1$,$p_c\rightarrow \infty$ are respectively defined as SPoC and MAC.

For the output $f$, it firstly is generalized by normalization through the $l_2$-normalization layer and dimensionality reduction through the full-connect layer. Finally, we concatenated the feature maps and trained it with the contrastive loss. In our model, we choosed the supervised contrastive loss as the contrastive loss function, while it is demonstrated that the gradient of supervised contrastive loss function encourages learning from hard positives and hard negatives. For $N$ images, the loss function takes the following form:
\begin{equation}
\begin{aligned}
    \mathcal{L} = - \sum_{i=1}^N log \frac{exp(z_i^T\cdot z^+/\tau)}{\sum_{j=1}^N exp(z_i^T \cdot z_j/\tau)},
\end{aligned}
\end{equation}
where $z_i$ denotes the anchor, $\tau\in R^+$ is s a scalar temperature parameter and $z^+$ is the number of the positives.
\section{Experiments}
\label{sec:experiments}
\subsection{Datasets}
The pet biometric dataset is collected by Ant Group, which consists of a training dataset, a validating dataset, and a testing dataset. The training set consisted of 6,000 dogs with 20,000 photographs of their nose prints, and each dog had at least two photographs. In the validating dataset, there are 2,000 pairs of dog nose print images (1,000 pair positives and 1,000 pair negatives), which were selected by a new dog print set that consists of 2,694 images. The testing dataset has 2,000 pairs of dog nose print images, which were selected by a new dog print set that consists of 4,000 images. The dataset format of training, validating, and testing can be seen in Table \ref{trainset} and Table \ref{val_test_set}.
\subsection{Experiment settings}
The training process is divided into two stages. Stage one is training stage, in which input images are resized to 60 $\times$ 60 pixels with probability 0.5 and then resized to 224 $\times$ 224 pixels. During stage two, we just finetune our model with  more image augmented strategies such as random brightness contrast and motion blur. To optimize our model, in stage one, we utilized the Adam optimizer with cosine weight decay scheduler for 30 epochs and tmax is set to 29. In stage two, we  finetune our mdoel for 20 epochs with smaller learning  rate, which is set to 3e-5 and tmax is 19. The embedding feature dimension is 512. In addition, we use EMA and AMP training strategies.

\begin{table}[!t]
\renewcommand{\arraystretch}{1.3}
\caption{Training dataset format.}
  \centering
\label{trainset}
\begin{tabular}{c|c}
\hline  
Dog ID & Nose Print Image\\
\hline 
0&x1FVNAVSRni...ARAD.jpg\\
0&--1WCesjS6C...ARAD.jpg\\
1&EJoldD9BQa-...ARAD.jpg\\
1&-27kQ7i7TMO...ARAD.jpg\\
...&...\\
5999&A*H4FeQZ5TNV...AAAQ.jpg\\
5999&dTUY37UVRhea...ARAD.jpg\\
5999&A*hxoTSL4Np6...AAAQ.jpg\\
\hline
\end{tabular}
\end{table}

\begin{table}[!t]

\renewcommand{\arraystretch}{1.3}
\caption{Validata and test dataset format.}
  \centering
\label{val_test_set}
\begin{tabular}{c|c}
\hline  
imageA & imageB\\
\hline 
A*p...AAQ.jpg&A*F...1AA.jpg\\
A*Y...AAQ.jpg&A*W...1AA.jpg\\
...&...\\
A*q...1AA.jpg&A*g...1AA.jpg\\
\hline
\end{tabular}
\end{table}

\subsection{Result of stage one}
In the first stage, the model is trained on traing dataset and tested on validate dataset. 
\begin{table}[!t]
\renewcommand{\arraystretch}{1.3}
\caption{The performance of our result varies from different backbone.}
  \centering
\label{backbone}
\begin{tabular}{c|c|c|c}
\hline  
backbone &image size &batch size& $AUC_{val}$\\
\hline 
EfficientNet-B1 NS&224&128 & 0.8479\\
EfficientNet-B2 NS&224&128 & 0.8460\\
EfficientNet-B3 NS&224&128 & 0.8489\\
EfficientNet-B4 NS&224&96 & 0.8540\\
EfficientNet-B5 NS&224&64 & 0.8599\\
Swin-B&224&88 & 0.8730\\
Swin-B&384&20 &0.8408\\
\hline
\end{tabular}
\end{table}
At first, we tried our model with different backbone networks, such as EfficientNet-B1-5 NS\cite{xie2020self}, Swin-B\cite{liu2021swin}. The results are shown in Table \ref{backbone}. We set this model with different batch sizes since CUDA memory is limited. We can see that large models perform better than small models. In table \ref{backbone}, the Swin-base model achieves the best performance.

\begin{table*}[!t]
\renewcommand{\arraystretch}{1.3}
\caption{Ablation results. ``w/o" represent without, ``entropy" represents a classification module, and ``pseudo 500" means that the most similar 500 pair validating images are labeled with the pseudo label, ``fc" represents fully connected layer, ``resize" means resize images with a probability, ``aug" represents image augmentation.}
  \centering
\label{stage1result}
\begin{tabular}{c|c|c}
\hline  
model &batch size& $AUC_{val}$\\
\hline 
Swin-base &32 & 0.8393\\
Swin-base &64 & $0.8477^{+0.0084}$\\
Swin-base + 10*cr&64 & $0.8529^{+0.0136}$\\
Swin-base w/o entropy & 88 & $0.8630^{+0.0237}$\\
Swin-base w/o entropy + pseudo 500 & 88 & $0.8700^{+0.0307}$ \\
Swin-base w/o entropy + pseudo 500 w/o fc & 88 & $0.8726^{+0.0333}$\\
Swin-base w/o entropy + pseudo 500 w/o fc resize & 88 & $0.8785^{+0.0392}$ \\
Swin-base w/o entropy + pseudo 500 w/o fc resize + aug& 88 & $0.8853^{+0.0460}$\\

\hline
\end{tabular}
\end{table*}

Secondly, based on the Swin-base model, we tried different tricks to improve the performance of our model, the results are shown in Table\ref{tricks}. The base model is a simple Swin-base model with a batch size of 32. Inspired by contrastive learning benefits from the large batch sizes, we enlarge the batch size from 32 to 64. We can see that the $AUC_{val}$ improved from 0.8393 to 0.8447. Since our model is developed by a CGD model, which was originally designed to solve image retrieval problems, we believe that the classification module would harm the performance of the dog nose print match task. We enlarge the weight of loss $\mathcal{L}_{cr}$, and the result shows that "Swin-base + 10*cr" is much higher than "Swin-base", which improved from 0.8477 to 0.8529. Taking a step further, we removed the classification module and enlarged the batch size, and the performance of "Swin-base w/o entropy" is achieved to 0.8630. Moreover, since the validating images are available during the training stage, we first trained a "Swin-base w/o entry" model, then infer the validating dataset and added the top 500 pair images to a new training stage. The result improved from 0.8630 to 0.8700. Next, the fully connected(FC) layer after GD may lose some information, we removed these FC layers in the infer stage. We can see that the performance of our model improved slightly.

\begin{figure}[t]
  \centering
   \includegraphics[width=0.8\linewidth]{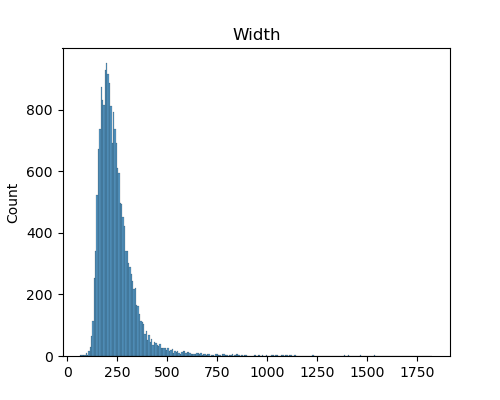}

   \caption{Image width frequency statistics of training dataset.}
   \label{train_img}
\end{figure}

\begin{figure}[t]
  \centering
   \includegraphics[width=0.8\linewidth]{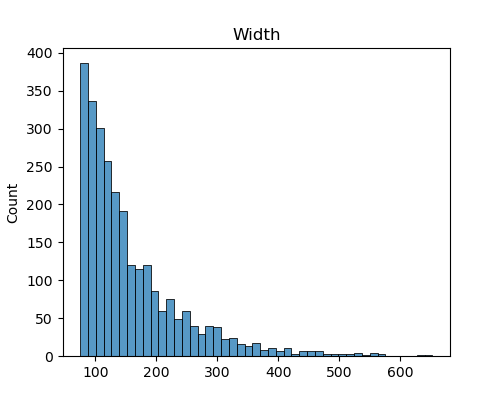}

   \caption{Image width frequency statistics of testing dataset.}
   \label{test_img}
\end{figure}
As shown in Figure \ref{train_img} and Figure \ref{test_img}, we can see that the width of training images is normally larger than 500, while the width of validating images is smaller than 200. There exists a big image size gap between the training dataset and validating dataset. So we resize the training images to one of [50, 60, 70, 80] with a probability of 0.45. The result shows that the resized image size model ``Swin-base w/o entropy + pseudo 500 w/o fc resize" improved from 0.8726 to 0.8785. At last, the original augmentation that existed in CGD may not be suitable for the dog nose print match task, so we removed the flip and crop augmentation and only keep ``resize" augmentation. The result also showed a significant improvement.

\begin{table}[!t]
\renewcommand{\arraystretch}{1.3}
\caption{Fusing multiple models.}
  \centering
\label{tricks}
\begin{tabular}{c|c|c}
\hline  
model &image size& $AUC_{val}$\\
\hline 
Swin-B& 224 & 0.8853\\
dm-nfnet-f3& 224 & 0.8915\\
Swin-B& 384 & 0.8900\\
EffNetV2-L& 224 & 0.8810\\
\hline
fusion &-& 0.9025\\ 
\hline
\end{tabular}
\end{table}

After applying these tricks in Swin-base, we also apply these tricks to some larger backbone models. Combined with the auto mix precision(AMP), some large backbones such as Swin384 can be trained by a larger batch size. In this setting, the dm-nfnet-f3 achieved 0.8915 on $AUC_{val}$, the Swin384, effv2-larger also achieved 0.8900 and 0.8810 on $AUC_{val}$, respectively. Last, we simply average the similarity of Swin-B(224), Swin-B(384), EffNetV2-L\cite{tan2021efficientnetv2}, and dm-nfnet-f3, and the result achieved to 0.9025.

\subsection{Result of stage two}

In the second stage, the effect of the test dataset inferred directly from the model trained in the first stage was relatively poor, and the online score was far lower than that in the first stage. Our online scores compared to other teams' online scores showed that their scores decreased even more. Our method outperformed many of the methods in the first stage, and we could see that our model performs better in generalization. In order to continued to enhance the performance, the following work involved data augmentation, without pseudo labels, test time augmentation, and models fusion.

Through the analysis of the data, we found that compared with the first stage data, there were more blurred and cropped images in the second stage. We combined a variety of augmentation schemes, first carried out two different scaling according to the probability, and cut into the same size of the image to facilitate input into the network. Subsequently image compression and blur operation, enhance image contrast and so on. The experimental results are shown in Table \ref{stage2_trick}.

In the first stage, part of the validata was used as pseudo labels. In the second stage, considering that the validata is a new dataset, it may have a bad influence to use the testset as pseudo labels.  Therefore the previous pseudo labels were eliminated and the models were to retrain 20 epochs on the basis of the previously trained model. It was discovered that efficientnet-b7 NS was considerably improved, while Swin-base was not improved. Our preliminary guess was that the Swin-Transformer need to be trained longer. Due to the limited time, we didn't attempt more backbones, but only removed the pseudo labels on efficientnet-b7 NS. The experimental results are shown in Table \ref{stage2_trick}.

\begin{table}[!t]
\renewcommand{\arraystretch}{1.3}
\caption{Tricks ablation experiments}
  \centering
\label{stage2_trick}
\begin{tabular}{c|c}
\hline  
tricks & $AUC_{test}$\\
\hline 
Efficientnet-B7 NS & 0.8450\\
Efficientnet-B7 NS + data augmentation & 0.8600\\
Efficientnet-B7 NS w/o pseudo labels & 0.8624\\
Efficientnet-B7 NS + TTA & 0.8460\\
final Efficientent-B7 NS & 0.8804\\
fusion four models with tricks & 0.8881 \\
\hline
\end{tabular}
\end{table}

With effective data augmentation, we tried a similar augmentation on the testset that using two scale scalings, random clipping, and random contrast augmentation. Image compression and blur were left out because some images were inherently fuzzy, and there were only slight differences between dog breeds. As it was a probabilistic selective augmentation, the amplitude of improvement of each experimental result was different, but each model could be steadily improved, as shown in Table \ref{stage2_trick}.

\section{Conclusion}
\label{sec:conclusion}
This paper, based on contrastive learning, proposes a new dog nose recognition network to extract the texture features of dog noses and then perform pairwise matching. Our approach is the first attempt to match dog species by contrastive learning. The method aims to obtain discriminative features and discriminate against unknown classes. Ablation experiments show that using only a single contrastive loss objective function is more stable than the combined performance ratio of classification loss and contrast loss, which further reflects the superiority of contrastive learning in handling unknown class matching tasks. In this competition, our model has a better performance in the real test scene, and the robustness is better than others. But the current test set AUC is 88.81\% and there is space for improvement. Directions that might be useful: try to extract features from key points of angle length instead of convolutional network features, try 3D reconstruction, try background culling, try to explore the effect of color changes on the results, try to explore the reason why the AUC decreased after alignment by key points.



{\small
\bibliographystyle{ieee_fullname}
\bibliography{egbib}

\begin{thebibliography}{10}\itemsep=-1pt

\bibitem{caya2021dog}
Meo~Vincent Caya, Emmanuel~D Arturo, and Chezjon~Q Bautista.
\newblock Dog identification system using nose print biometrics.
\newblock In {\em 2021 IEEE 13th International Conference on Humanoid,
  Nanotechnology, Information Technology, Communication and Control,
  Environment, and Management (HNICEM)}, pages 1--6. IEEE, 2021.

\bibitem{chakraborty2020investigation}
Shoubhik Chakraborty, Kannan Karthik, and Santanu Banik.
\newblock Investigation on the muzzle of a pig as a biometric for breed
  identification.
\newblock In {\em Proceedings of 3rd International Conference on Computer
  Vision and Image Processing}, pages 71--83. Springer, 2020.

\bibitem{chen2016locality}
Yu-Chen Chen, Shintami~C Hidayati, Wen-Huang Cheng, Min-Chun Hu, and Kai-Lung
  Hua.
\newblock Locality constrained sparse representation for cat recognition.
\newblock In {\em International Conference on Multimedia Modeling}, pages
  140--151. Springer, 2016.

\bibitem{deb2018face}
Debayan Deb, Susan Wiper, Sixue Gong, Yichun Shi, Cori Tymoszek, Alison
  Fletcher, and Anil~K Jain.
\newblock Face recognition: Primates in the wild.
\newblock In {\em 2018 IEEE 9th International Conference on Biometrics Theory,
  Applications and Systems (BTAS)}, pages 1--10. IEEE, 2018.

\bibitem{khosla2020supervised}
Prannay Khosla, Piotr Teterwak, Chen Wang, Aaron Sarna, Yonglong Tian, Phillip
  Isola, Aaron Maschinot, Ce Liu, and Dilip Krishnan.
\newblock Supervised contrastive learning.
\newblock {\em Advances in Neural Information Processing Systems},
  33:18661--18673, 2020.

\bibitem{kumar2016face}
Santosh Kumar, Shrikant Tiwari, and Sanjay~Kumar Singh.
\newblock Face recognition of cattle: can it be done?
\newblock {\em Proceedings of the National Academy of Sciences, India Section
  A: Physical Sciences}, 86(2):137--148, 2016.

\bibitem{liu2021swin}
Ze Liu, Yutong Lin, Yue Cao, Han Hu, Yixuan Wei, Zheng Zhang, Stephen Lin, and
  Baining Guo.
\newblock Swin transformer: Hierarchical vision transformer using shifted
  windows.
\newblock In {\em Proceedings of the IEEE/CVF International Conference on
  Computer Vision}, pages 10012--10022, 2021.

\bibitem{tan2021efficientnetv2}
Mingxing Tan and Quoc Le.
\newblock Efficientnetv2: Smaller models and faster training.
\newblock In {\em International Conference on Machine Learning}, pages
  10096--10106. PMLR, 2021.

\bibitem{wang2019learning}
Hongnian Wang, Han Su, Peng Chen, Rong Hou, Zhihe Zhang, and Weiyi Xie.
\newblock Learning deep features for giant panda gender classification using
  face images.
\newblock In {\em Proceedings of the IEEE/CVF International Conference on
  Computer Vision Workshops}, pages 0--0, 2019.

\bibitem{xie2020self}
Qizhe Xie, Minh-Thang Luong, Eduard Hovy, and Quoc~V Le.
\newblock Self-training with noisy student improves imagenet classification.
\newblock In {\em Proceedings of the IEEE/CVF conference on computer vision and
  pattern recognition}, pages 10687--10698, 2020.

\end{thebibliography}
}

\end{document}